\documentclass[a4paper]{mcs12}

\usepackage{tabularx} 
\usepackage{graphicx} 
\usepackage[intlimits]{amsmath}
\usepackage{amssymb} 
\usepackage{times}
\usepackage{soul} 
\usepackage[square,numbers,sort&compress]{natbib} 
\usepackage[labelfont=bf,font=normalsize]{caption}  
\usepackage{epstopdf} 
\usepackage{hyperref} 
\hypersetup{
    colorlinks=true,
    citecolor = blue,
    linkcolor=blue,
    urlcolor = blue
}

\begin{document}

\title{Accurate ignition detection of solid fuel particles using machine learning}

\authors{T. Li*, Z. Liang*, A. Dreizler* and B. Böhm*}

\corrauthoremail{tao.li@rsm.tu-darmstadt.de}

\address{*Technical University of Darmstadt, Department of Mechanical Engineering, Reactive Flows and Diagnostics, Otto-Berndt-Straße 3, 64287 Darmstadt, Germany}

\begin{abstract} 
In the present work, accurate determination of single-particle ignition is focused on using high-speed optical diagnostics combined with machine learning approaches.~Ignition of individual particles in a laminar flow reactor are visualized by simultaneous 10\,kHz OH-LIF and DBI measurements.~Two coal particle sizes of 90\,-\,125\,\textmu m and 160\,-\,200\,\textmu m are investigated in conventional air and oxy-fuel conditions with increasing oxygen concentrations.~Ignition delay times are first evaluated with threshold methods, revealing obvious deviations compared to the ground truth detected by the human eye.~Then, residual networks (ResNet) and feature pyramidal networks (FPN) are trained on the ground truth and applied to predict the ignition time.~Both networks are capable of detecting ignition with significantly higher accuracy and precision.~Besides, influences of input data and depth of networks on the prediction performance of a trained model are examined.~The current study shows that the hierarchical feature extraction of the convolutions networks clearly facilitates data evaluation for high-speed optical measurements and could be transferred to other solid fuel experiments with similar boundary conditions.

\end{abstract}

\section{Introduction}
\label{sec:intro}
\vspace{1mm}
Particle ignition is an essential stage for the flame stability of a pulverized fuel stream, which has been a research topic since years \cite{Howard.1967Symp.(Int.)Combust.,Essenhigh.1989Combust.Flame, Annamalai.1993Prog.EnergyCombust.Sci.,Du.1994Combust.Flame}.~In general, particle ignition can be classified into two modes: homogeneous and heterogeneous ignition.~In homogeneous ignition, volatile matters, including hydrogen (H$_2$), hydrocarbons (C$_x$H$_y$), and tars, are released from particles upon increasing temperatures.~These fuel gases, mixing with oxidizer, ignite above the flammable temperature and mixture fraction limits, resulting in a gas-phase flame in the vicinity of the particle.~In heterogeneous ignition, oxidizers approach the particle surface followed by direct surface reactions.~The ignition mode can be influenced by coal rank, particle size, heating rate, and gas composition.~

On the one hand, particle surface temperature is a widely accepted parameter for determining heterogeneous ignition.~A well-calibrated two- or three-color pyrometer can provide conclusive information about particle temperature and heterogeneous combustion \cite{Levendis.1992,Bejarano.2008Combust.Flame,Yuan.2016Fuel}.~On the other hand, indicators and corresponding measurement techniques are rather diverse for homogeneous combustion.~Here, particle temperature is still an essential parameter in experiments, see for example \cite{Khatami.2011Combust.Flame,Levendis.2011Combust.Flame,Khatami.2012Combust.Flame,Maffei.2013Combust.Flame,Khatami.2015Combust.Flame,Lei.2017, Vorobiev.2020Combust.Sci.Technol.}.~However, the pyrometric temperature closely relates to hot soot and char particles.~Recent advancements in high-speed imaging enables tracking the particle combustion history with sufficient temporal and spatial resolutions \cite{Khatami.2012Combust.Flame,Cai.2015Fuel,Khatami.2016Combust.Flame,AdewaleAdeosun.2018}.~These investigations define the gas-phase ignition on the first visible broad-band emission signal recorded by high-speed cameras in the kHz range.~Methylidyne (CH) chemiluminescence was also used as an ignition indicator, which can be imaged in a time-integrated \cite{Molina.2007Proc.Combust.Inst.} or single-shot \cite{Yuan.2016Fuel} manner.~For single-shot measurements, intensified cameras are required due to the weak CH* chemiluminescence signals.~In addition, soot particles were imaged by an intensified camera to derive the homogeneous ignition time \cite{Shaddix.2009Proc.Combust.Inst.}.~Since the CH* correlates with the heat release in the gas phase, it is a better indicator than the black-body radiation of soot particles.~Unfortunately, CH* emission (at about 430\,nm) spectrally overlaps with the broad-band black-body emission.  

Besides CH radicals, the hydroxyl (OH) radical has been used as an important flame marker, which abundantly exists in the reaction zone and the burned gas.~The first planar laser-induced fluorescence of OH radicals (OH-LIF) on the single-particle level was reported in \cite{Koser.2015Appl.Phys.B}.~The homogeneous ignition process of individual particles was visualized at 10\,kHz with a spatial resolution of approximately 100\,\textmu m.~The technique enabled the evaluation of single-particle volatile flame structures with different particle sizes and atmospheres \cite{Koser.2017Proc.Combust.Inst.}.~Further experiments, supplemented by simultaneous flame luminosity and diffused backlight-illumination (DBI), indicated the distinctive appearance of OH-LIF and luminosity signals at the onset of ignition, suggesting OH-LIF as a favorable diagnostic approach for homogeneous ignition detection \cite{Koser.2019Proc.Combust.Inst.}. 

Accurate ignition detection depends on combustion conditions, experimental techniques, as well as the definition of the ignition event \cite{ZHANG.1994Fuel}.~The onset of ignition is usually defined based on a certain intensity threshold \cite{Molina.2007Proc.Combust.Inst.,Koser.2019Proc.Combust.Inst.,Yuan.2016Fuel} above the background level or the evaluation of signal topology \cite{Li.2021Fuel,Shaddix.2009Proc.Combust.Inst.}.~Parameter modification in the image analysis is unavoidable if a different particle or particle size is investigated, or another detection system is used.~This hurdle can be tackled by deep learning for a better feature extraction.~Deep learning approaches are capable of learning image features and making predictions after training on a data set.~Recently, convolution neural networks (CNN) have been applied in experimental studies to predict velocity fields \cite{Barwey.2022} and 3D flame structures \cite{Huang.2019J.FluidMech.}.~In the present work, high-speed multi-parameter laser diagnostics are assisted by deep learning approaches, providing accurate ignition delay time based on object classification and detection architectures.

Considering its importance and challenges, the current study emphasizes on the accurate ignition detection in solid fuel combustion, which was not fully addressed in the previous experiment.~Homogeneous ignition of high-volatile solid fuels at realistic heating rates is particularly targeted.~For this purpose, simultaneous OH-LIF and DBI measurements were applied for single particles burning in a laminar flow reactor (LFR), producing a comprehensive database.~Homogeneous ignition is evaluated for two particle sizes under air and oxy-fuel atmospheres.~Two main deep learning architectures, namely residual network (ResNet) and feature pyramidal network (FPN) are implemented.~The influence of training data, network depth, and pre-training on the capability for ignition prediction is carefully examined.  

\section{Experiments}
\label{sec:exp}
\vspace{1mm}
Multi-parameter optical diagnostics were employed to investigate the fundamental processes of single coal particles during ignition and volatile combustion in laminar flows.~The experimental details were reported in the previous work \cite{Li.2021Fuel} and will be briefly introduced here.~An in-house laminar flow reactor was designed to generate desired gas atmospheres simulating temperatures and species concentrations in realistic conditions.~Premixed lean methane (CH$_4$) flat flames were stabilized on the surface of a ceramic matrix.~By operating inlet gas mixtures of CH$_2$/O$_2$/N$_2$ or CH$_2$/O$_2$/CO$_2$, conventional (AIR) and oxy-fuel (OXY) atmospheres were generated in the exhaust gas with homogeneous temperature and velocity fields.~These conditions were denoted as AIR10/20/30/40 and OXY20/30/40, with the number indicating the volumetric oxygen concentrations of the exhaust gas.~Micrometer-sized bituminous coal particles were seeded individually into the burner by carrier gases having the same molecular composition and velocity as the flat flame inlet.   

\begin{figure}[h!]
\centering
\includegraphics[width=70mm]{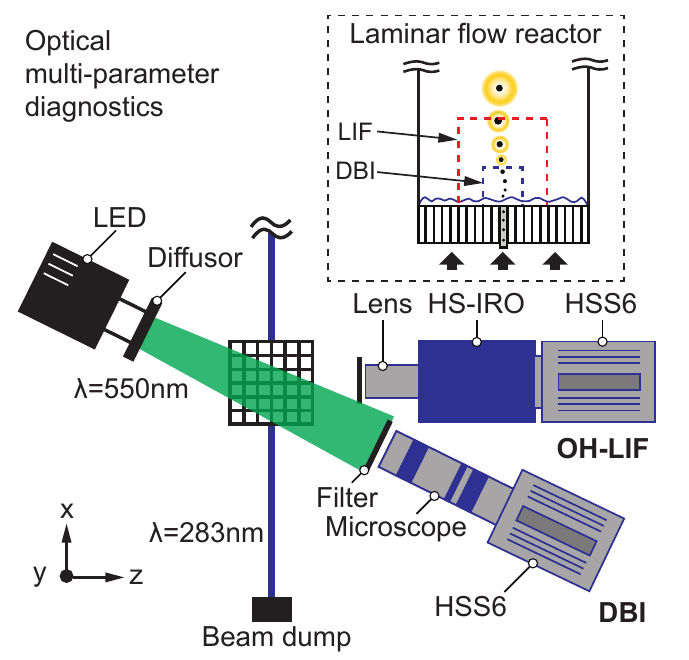}
\caption{A schematic experimental layout including optical diagnostics and the laminar flow reactor.}
\label{Fig_Setup_SPC_ML}
\end{figure}

Figure \ref{Fig_Setup_SPC_ML} illustrates the multi-parameter optical measurement and a sketch of the burner.~Gas-phase volatile flames were visualized by two-dimensional high-speed OH-LIF at 10\,kHz.~OH-LIF signals were excited at 283.01\,nm by a dye laser system and collected by a CMOS camera (HSS6, LaVision) coupled with a lens-coupled intensifier (HS-IRO, LaVision).~A narrow band-pass filter and a short gate time were applied to eliminate interfering thermal radiation and chemiluminescence.~Particle ignition time can be determined by tracking the temporal variation of OH-LIF signal topology \cite{Li.2021Fuel}.~Besides, particles positions and velocities are simultaneously detected by 10\,kHz DBI measurements.~A high-power LED illuminated coal particles at 550\,nm and projected particle shadows were imaged onto another CMOS camera (HSS6, LaVision).~This camera was equipped with a long-distance microscope to enhance the spatial resolution estimated as approximately 20\,\textmu m at the burner center position \cite{Li.2022Proc.Combust.Inst.b}.~The DBI technique allowed for an accurate determination of particle size, shape, and velocities \cite{Li.2021Fuel,Li.2022Proc.Combust.Inst.b}.~As illustrated in Fig.\ref{Fig_Setup_SPC_ML}, the difference in resolutions led to a smaller field of view for DBI (i.e. $11 (\textrm{height}) \times 5 (\textrm{width}) $\,mm$^2$) than for OH-LIF (i.e. $19 \times 19$\,mm$^2$).~Since the homogeneous ignition of single particles occurs within a few millimeters above the burner, the ignition process can be fully captured by both imaging techniques.~More details regarding experimental methodology are referred to \cite{Li.2021Fuel}. 

\section{Data evaluation}
\vspace{1mm}
Colombian high-volatile bituminous coal particles sieved to two size distributions of 90\,-125\,\textmu m and 160\,-200\,\textmu m were investigated, which are referred as particles A and B.~With an approximately 10$^5$\,K/s particle heating rate \cite{Li.2021Fuel,Li.2022Proc.Combust.Inst.b}, homogeneous ignition of released volatiles dominates the particle ignition mode.~The ignition process was captured in the so-called single-particle event, in which an individual particle moves through the probe volume without interacting with other particles.~In total, 1006 events and 512 events were detected for particles A and B, respectively.~Due to the larger diameter, particle B was more difficult to seed through the 0.8\,mm injection tube, resulting in lower probabilities of particle detection.

\begin{figure*}[h!]
\centering
\includegraphics[width=150mm]{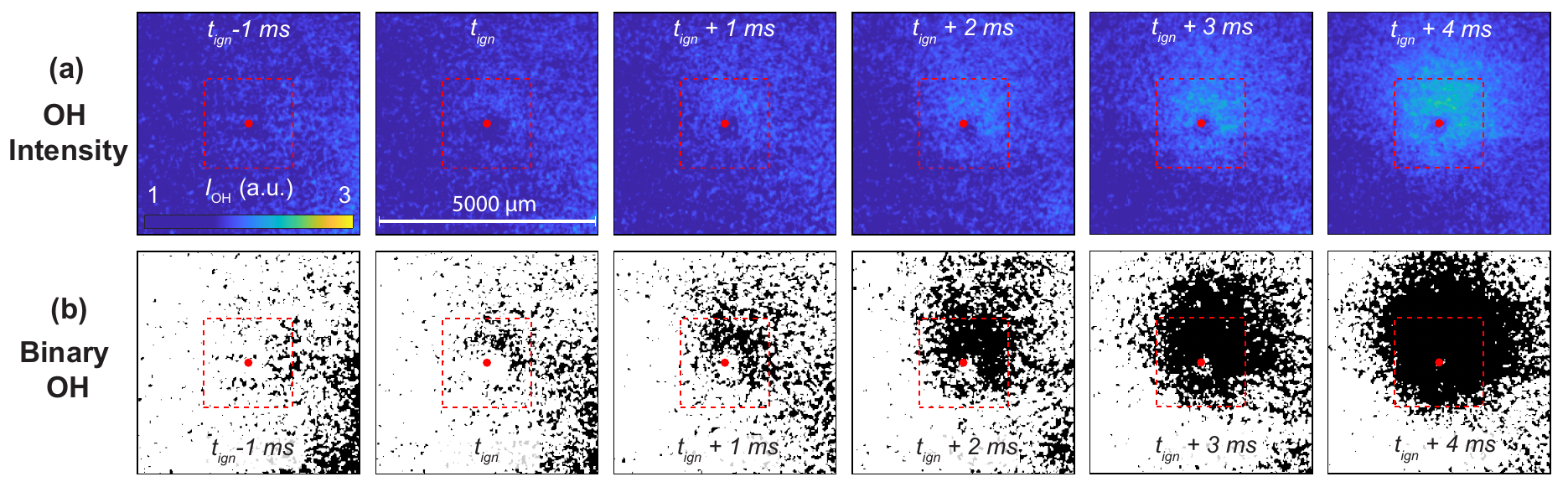}
\caption{A time-resolved sequence of particle ignition with $t_\textrm{ign}$ given by the ground truth (manual labeling).~(a) OH-LIF raw images.~(b) binary OH-LIF images.}
\label{Fig_Ignition_Sequence}
\end{figure*}

An example of a single-particle event is presented in Fig.\,\ref{Fig_Ignition_Sequence}, temporally conditioned on the ignition instance at $t_ \textrm{ign}$ (every tenth images shown).~Figure\,\ref{Fig_Ignition_Sequence}(a) shows the temporal evolution of OH-LIF intensities normalized on the homogeneous background signal.~By applying a constant intensity threshold $I_\textrm{th}$ of 1.2 \cite{Li.2021Fuel}, corresponding binary images are shown in Fig.\ref{Fig_Ignition_Sequence}(b).~In the previously proposed signal and structure (SAS) analysis \cite{Li.2021Fuel}, the onset of ignition is defined if the normalized intensity and the connected area exceed respective thresholds.~The center points of particles (red solid dots) are shot-by-shot determined from the DBI data.~Since all cameras are spatially mapped based on an accurate target calibration, DBI center points reveal as good references to locate particles in OH-LIF images. 

To assess the accuracy of different methods for ignition detection, a ground truth of ignition delay time needs to be defined for each particle event.~For this purpose, the ignition frame of all 1518 events is manually labeled the same person, who is instructed to select the first OH-LIF image, which contains recognizable flame structures against the background.~The performance for predicting single-particle ignition can be evaluated using the ignition time difference (ITD): 

\begin{equation}
\begin{aligned}
    \textrm{ITD}_\textrm{detector} = t_\textrm{i,detector} -t_\textrm{i,gt}.  \\
\end{aligned} 
\label{ITD}
\end{equation}

\noindent Here, $t_\textrm{i,detector}$ and $t_\textrm{i,gt}$ indicate the ignition delay time determined by a detector and ground truth, respectively.~If ITD $>$ 0, ignition time is over-predicted, otherwise ignition time is under-predicted.~A statistical analysis was performed for each particle size to compare the accuracy and precision of each detector or data processing method.

Two different deep learning networks are implemented, which belong to object classification (i.e, residual networks, ResNet) and object detection (feature pyramidal networks, FPN) in the context of computer vision.~Image classification works on an entire image or an image segment and classifies the image into a category, whereas object detection specifies multiple objects within an image into categories (e.g. classification) with their locations represented by bounding boxes (e.g. bounding box regression).~For brevity, the architecture of deep learning networks are not elaborated here, and more details of related work are referred to \cite{He_2016_CVPR} for ResNet, \cite{Girshick_2014_CVPR} for R-CNN, \cite{Girshick_2015_ICCV} for Fast R-CNN, \cite{NIPS2015_14bfa6bb} for Faster R-CNN, and \cite{Lin_2017_CVPR} for FPN.

\section{Results and Discussions}
\vspace{1mm}
\subsection{Ignition detected by SAS}
\begin{figure*}[h!]
\centering
\includegraphics[width=140mm]{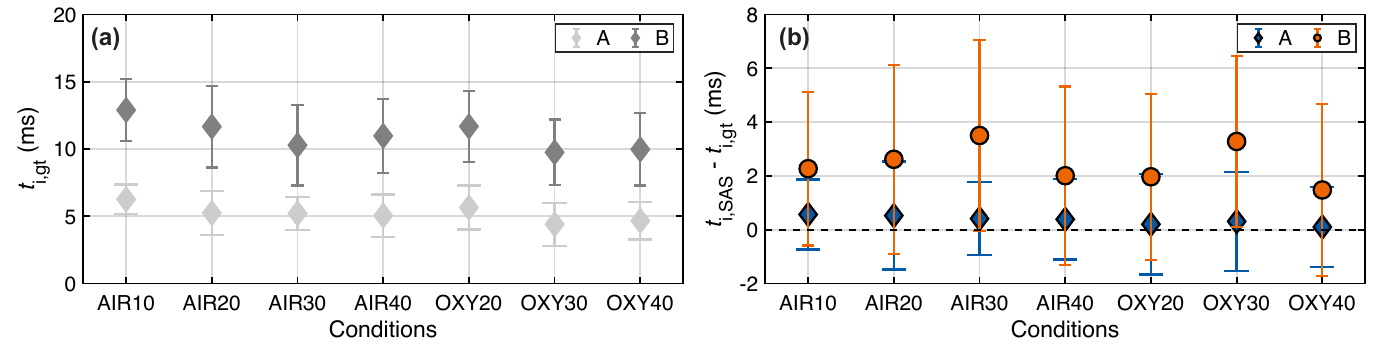}
\caption{Comparison of ignition delay times by the SAS method $t_\textrm{i,SAS}$ and the manual label $t_\textrm{i,gt}$ for two particle sizes A and B in seven atmospheres.}
\label{Fig_tign_SAS_GT_Compare}
\end{figure*}

\vspace{1mm}
Figure\,\ref{Fig_tign_SAS_GT_Compare}(a) shows the ignition delay time determined by ground truth ($t_\textrm{i,gt}$) for particles A and B under seven investigated atmospheres.~Here, symbols and error bars indicate the mean value ($\mu$) and $\pm$ one standard deviation ($\sigma$), respectively.~The ignition delay time is referenced to the flat flame position ($y$\,$\approx$\,1.5\,mm) where particles start to heat up by the hot exhaust gas.~Basically, the overall trend over particle diameters and atmospheres is very similar to $t_\textrm{i,SAS}$ in \cite{Li.2021Fuel}.~Effects of particle size, oxygen concentration, and CO$_2$ replacement can be observed, and more details can be found in \cite{Li.2021Fuel}.~Differences between $t_\textrm{i,gt}$ and $t_\textrm{i,SAS}$ are observed.~The over-prediction of ignition times by the SAS analysis can be quantified by calculating the ignition time difference ITD for each single-particle event, as shown in Figure\,\ref{Fig_tign_SAS_GT_Compare}(b).~Overall good agreement is archived by the SAS analysis for the small particles A, whereas some discrepancies can be identified for the large particle B.~Specifically, the SAS analysis over-predicts the ignition delay time of particles B by 2\,-\,4\,ms in all atmospheres, which is presumably caused by the improper selection of intensity and area thresholds.~This challenge motivated the application of machine learning for the more generalized feature detection of ignited particles.

\subsection{Ignition detected by ResNets}
\vspace{1mm}
In machine learning approaches, only 35\% of the entire events are used to train ResNet models \cite{He_2016_CVPR}, which corresponds to 462 particle events.~The remaining 65\% (1056 events) are testing data, allowing for sufficient statistical comparisons with ground truth and the performance evaluation of different architectures.~However, if the 35\% data were further divided into training data and validation data, the number of training data would be insufficient to obtain converge networks.~Therefore, the $k$-folds cross-validation approach is consistently used to train object classification networks.~With $k$-folds cross-validation, the entire training set is split into $k$ sets.~One by one, a set is selected for validation, and the $k-1$ other sets are combined into the corresponding training set.~5-folds cross-validation is used, and 10 epochs are trained for each fold.~A predicted probability higher than 50\% is classified as an ignited particle in testing data.

\begin{figure*}[h!]
\centering
\includegraphics[width=140mm]{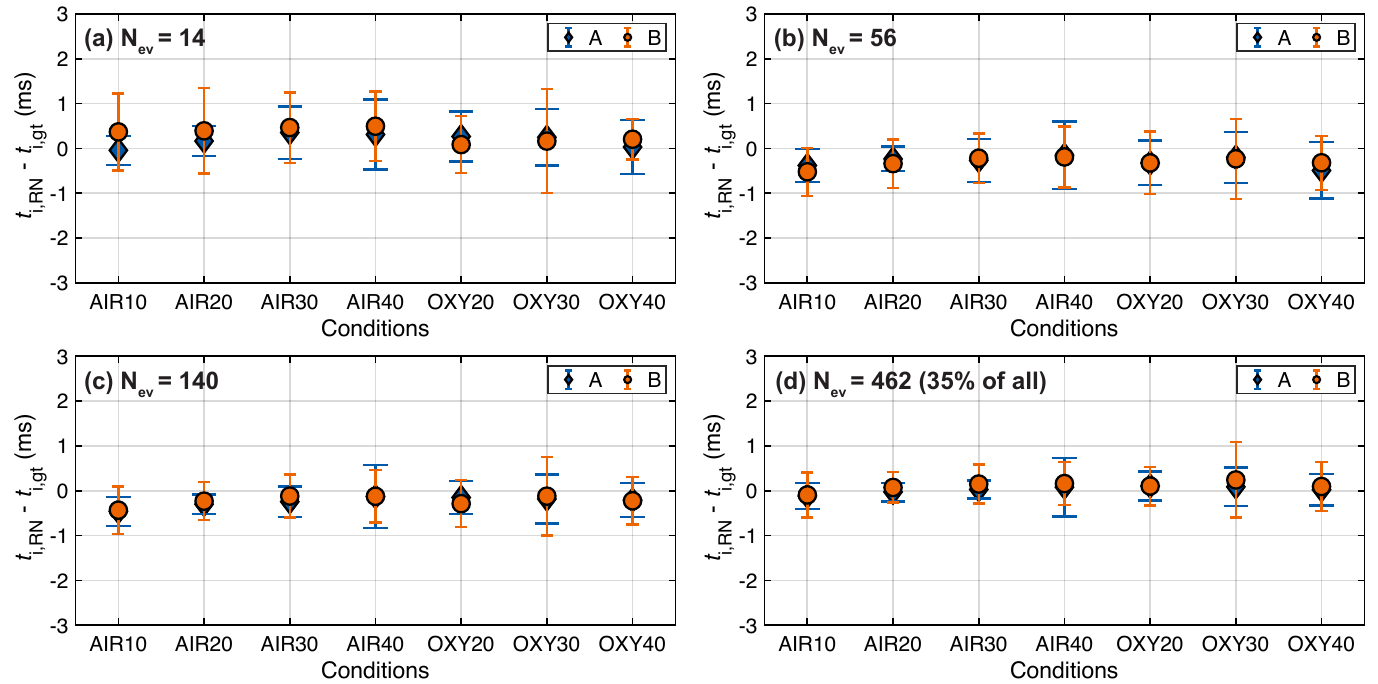}
\caption{Ignition time difference $t_\textrm{i,RN}$\,-\,$t_\textrm{i,gt}$ by using ResNet-18 with the amount of particle events (a) $N_\textrm{ev}$\,=\,14, (b) $N_\textrm{ev}$\,=\,56, (c) $N_\textrm{ev}$\,=\,140, and (d) $N_\textrm{ev}$\,=\,462.}
\label{Fig_delta_tign_resnet18_Nev_effect}
\end{figure*}

Several ResNet-18 models are trained with an increasing number of particle events, which are equally selected from different operating conditions, in order to examine the influence of training data.~Figure\,\ref{Fig_delta_tign_resnet18_Nev_effect} shows the ignition time difference between ground truth $t_\textrm{t,gt}$ and $t_\textrm{i,RN}$ predicted by ResNet-18 which is trained with increasing particle events $N_\textrm{ev}$ = 14, 56, 140, and 462.~They correspond to 1, 4, 10, and 33 (equivalent) particle events selected from each atmosphere and particle size.
~Note that new data are always added into the training by retaining the existing data, e.g., the original 14 events are included in the 56 events and so on.~For a reasonable comparison, 1056 particle events are consistently used for the prediction purpose.~Generally, ResNet-18 achieves a higher precision with less scattered results than the SAS methods.

With more data fed into the training process, mean values of $t_\textrm{i,RN}$\,-\,$t_\textrm{i,gt}$ (ITD) approach 0\,ms indicating an continuously improving accuracy in predicting the ground truth.~With a small amount of data, e.g. $N_\textrm{ev} \le 140$, ITD distributions shift in either negative or positive side.~It implies that the networks are still fragile to the newly added training data, as weight and bias parameters are not sufficiently trained with a limited number of events.~With $N_\textrm{ev}$ increased to 462, the predictive accuracy becomes equally high for particle A and B.~Furthermore, error bars in Fig.\,\ref{Fig_delta_tign_resnet18_Nev_effect} constantly narrow with increasing data involved in the training process.~It is obvious that the number of images used for training improves the precision of ignition prediction.~For ignition detection, 35\% of data for training is proposed as a reasonable compromise between the performance of trained networks and the expanse of manual labeling whenever a new data set is under evaluation.

\subsection{Ignition detected by FPN}
\begin{figure*}[h!]
\centering
\includegraphics[width=140mm]{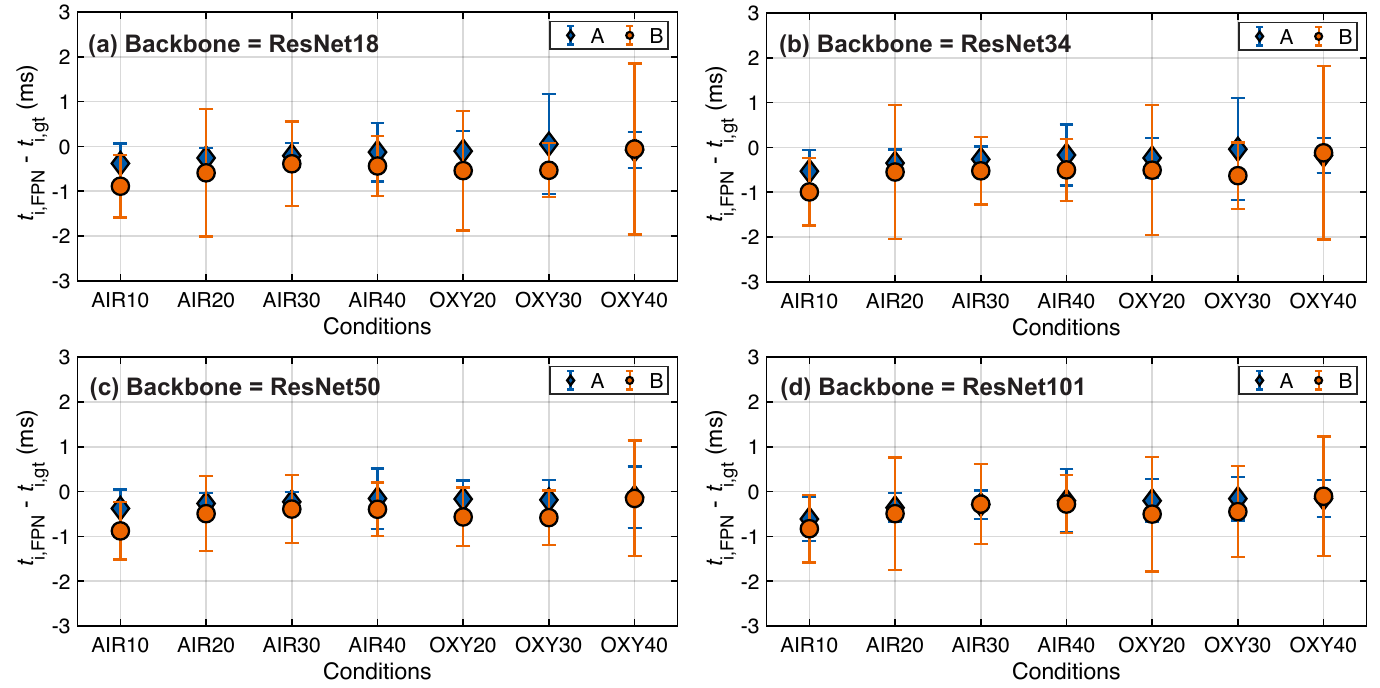}
\caption{Ignition time difference $t_\textrm{i,FPN}$\,-\,$t_\textrm{i,gt}$ using different ResNet models in the bottom-up pathway of FPN networks.}
\label{Fig_sigma_mu_FPN}
\end{figure*}

\vspace{1mm}
For FPN \cite{Lin_2017_CVPR}, training loss converges, and 5 epochs are trained to reduce the computational cost.~A particle will be identified as $ignited$ for the final classification if its predicted probability is higher than 50\%.~Figure\,\ref{Fig_sigma_mu_FPN} compares predictive results of FPN by implementing ResNets with an increasing depth in the bottom-up pathway.~All models are trained with 462 particle events.~The ignition time difference $t_\textrm{i,FPN}$\,-\,$t_\textrm{i,gt}$ is represented by its mean and standard deviation.~An overall good agreement between FPN prediction and ground truth is observed, which is independent on atmosphere but on particle size.~Compared to the SAS method in Fig.\ref{Fig_tign_SAS_GT_Compare}, evident improvements are achieved in predictive accuracy and precision.~However, compared to ResNet-18 in Fig\,\ref{Fig_delta_tign_resnet18_Nev_effect}(d), the ITD scatters more in FPN by using the same training data.~In addition, it can be noted that increasing backbone convolutions layers narrows the error bars and improves the prediction precision.~But no further improvements can be noticed comparing backbone ResNet-101 with ResNet-50.

\section{Conclusions}
\begin{figure}[h!]
\centering
\includegraphics[width=70mm]{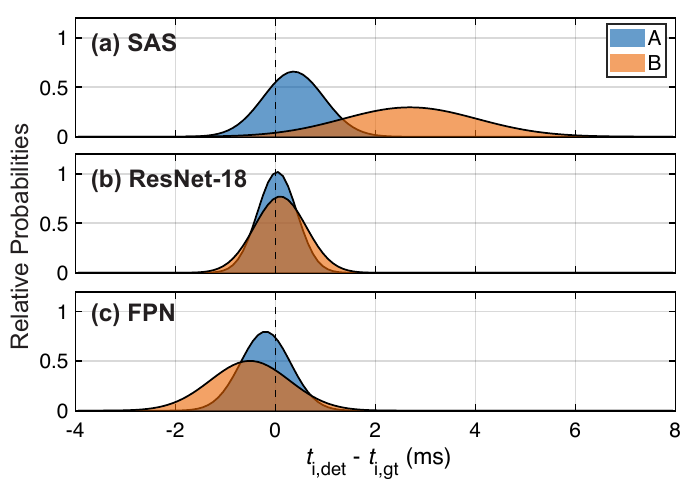}
\caption{Relative probability distributions of ignition time differences $t_\textrm{i,det}$\,-\,$t_\textrm{i,gt}$, in which $t_\textrm{i,det}$ represents predicted ignition delay times by different approaches.}
\label{Fig_tign_all_methods_compare}
\end{figure}

\vspace{1mm}
The current work presents an experimental investigation of single particle ignition using high-speed optical diagnostics in a laminar flow reactor.~Homogeneous ignition is visualized by 10\,kHz OH-LIF measurements with the simultaneously detected particle location in DBI measurements.~Accurate detection of ignition delay times is focused on with both conventional threshold methods and advanced machine learning approaches.~The prediction performance of different approaches for ignition detection are conclusively compared in Fig.\,\ref{Fig_tign_all_methods_compare}.~The mean $\mu$ and standard deviation $\sigma$ of ITD are evaluated by including all atmospheres and further used to generate normal distributions $\mathcal{N}(\mu$, $\sigma$) approximating the overall prediction performance for particles A and B.~The previously introduced SAS method provides satisfactory predictions for small particles but substantially over-estimates large particles' ignition.~It is because the accuracy and precision inherently relate with the threshold selection.~Owing to the changing characteristics of OH-LIF signals with particles and atmospheres, fixed thresholds for area and intensities might induce errors and restrict the detection quality.~Sensitivity analyses (not shown) clearly indicates the distinctive signals, especially the intensity levels at the onset of ignition, which makes conventional threshold methods incapable for thus a task.~Although optimum thresholds can be statistically obtained by minimizing the difference between prediction and ground truth, this method is not really viable when dealing with a new set of data, owing to the in-prior knowledge required for optimal thresholding.~To avoid the difficulties in optimizing algorithm parameters, convolutions networks with hierarchical feature extraction are implemented. 

Figure\,\ref{Fig_tign_all_methods_compare}(b) and (c) show the results provided by the best ResNet model (i.e.~pre-trained ResNet-18 fine-tuned with 462 events) and the best FPN model (i.e. backbone ResNet-50 trained with 462 events), respectively.~Evidently, enhanced quality of ignition detection is achieved, which is superior to the conventional processing approach investigated in this study.~This can be explained by the feature recognition over different scales, which is inherently included in the hierarchy of convolutional layers in their architectures.~As a result, ResNet-18 achieves the most accurate and precise ignition delay time compared with ground truth.~More complex networks such as FPN promote no further improvements but slight under-estimation of ignition delay times.~However, training ResNet-18 involves an additional processing step of RoI extraction from the particle center.~Despite FPN models also require pre-estimated particle positions but is able to work on an entire image.~Although these models are heavier and need a longer training time, they could be further developed to detect multi-particle ignition in the future.~Regarding the simple features of ignited coal particles, it can be concluded that both object classification and detection approaches of machine learning are valuable for solid fuel combustion analysis.~Residual and feature pyramidal networks are appropriate architectures for such evaluation tasks and have the potential to be transferred to other experimental investigations.

\section{Acknowledgements}
This work was founded by the Hessian Ministry of Higher Education, Research, Science and the Arts - cluster project Clean Circles.

\section{References}
{\bibliographystyle{mcs12}
\setlength{\bibsep}{0.5mm}
\def\section*#1{}
\bibliography{ref}}

\end{document}